\documentclass{article}

\usepackage[preprint]{neurips_2021}

\usepackage[utf8]{inputenc} 
\usepackage[T1]{fontenc}    
\usepackage{hyperref}       
\usepackage{url}            
\usepackage{booktabs}       
\usepackage{amsfonts}       
\usepackage{nicefrac}       
\usepackage{microtype}      
\usepackage{xcolor}         

\title{Neural Density Estimation and Uncertainty Quantification for Laser Induced Breakdown Spectroscopy Spectra}

\author{
  Katiana Kontolati \\
  Johns Hopkins University\\
  Baltimore, MD 21218 \\
  \texttt{kontolati@jhu.edu} \\
  \And
  Natalie Klein \\
  Los Alamos National Laboratory\\
  Los Alamos, NM 87545 \\
  \texttt{neklein@lanl.gov} \\
  \AND
  Nishant Panda \\
  Los Alamos National Laboratory\\
  Los Alamos, NM 87545 \\
  \texttt{nishpan@lanl.gov} \\
  \And
  Diane Oyen \\
  Los Alamos National Laboratory\\
  Los Alamos, NM 87545 \\
  \texttt{doyen@lanl.gov} \\
}

\begin{document}

\maketitle

\begin{abstract}
  Constructing probability densities for inference in high-dimensional spectral data is often intractable. In this work, we use normalizing flows on structured spectral latent spaces to estimate such densities, enabling downstream inference tasks. In addition, we evaluate a method for uncertainty quantification when predicting unobserved state vectors associated with each spectrum. We demonstrate the capability of this approach on laser-induced breakdown spectroscopy data collected by the ChemCam instrument on the Mars rover Curiosity. Using our approach, we are able to generate realistic spectral samples and to accurately predict state vectors with associated well-calibrated uncertainties. We anticipate that this methodology will enable efficient probabilistic modeling of spectral data, leading to potential advances in several areas, including  out-of-distribution detection and sensitivity analysis.
\end{abstract}

\section{Introduction}

The ChemCam instrument on the Mars rover Curiosity uses laser-induced breakdown spectroscopy (LIBS), a type of an atomic emission spectroscopy, to remotely analyze Martian rocks \citep{wiens2012chemcam}. Spectral information is used to extract the qualitative and quantitative chemical content (QQCC) of a material sample, where the QQCC can be seen as an unobserved state vector representing the sample. Both linear and nonlinear supervised learning techniques have been applied for mapping LIBS spectra to QQCC with good accuracy \citep{forni2013independent, boucher2015study, castorena2021deep}. In this work, we build upon these efforts by proposing a framework for constructing the probability density function (PDF) of LIBS spectra. In addition, unlike previous methods which produce point estimates of QQCC, we propose and evaluate a method for uncertainty quantification (UQ) on point predictions of QQCC. 

Many real-world spectra, including LIBS spectra, are characterized by a large number of features, complicating the construction of the data PDF due to the curse of dimensionality. We develop a novel framework for constructing low-dimensional PDFs suitable for downstream inference (e.g., sampling, density estimation, outlier detection, or unsupervised representation learning) using state-of-the-art neural density estimators with normalizing flows (NF) on spectral latent spaces. This framework allows us to generate realistic spectral samples on a reduced space and using an inverse-transformation project them back to the physically interpretable space. Furthermore, we propose a bootstrapping approach for quantifying uncertainty in predictions of unobserved state vectors corresponding to each spectrum. We demonstrate the capabilities of the proposed approach to construct the PDF of a LIBS spectral data set, and to learn a mapping to the known QQCC with uncertainty. The validated framework can be then employed for QQCC prediction and direct UQ of novel samples, such as artificial samples generated by the NF model as well as spectra collected directly on Mars. 

To the best of our knowledge, this work is the first time normalizing flows are constructed on spectral latent spaces and can be readily employed for any kind of spectroscopy data. We show that the proposed framework provides a straightforward way to perform downstream inference tasks and direct UQ for high-dimensional spectral data. 

\section{Methods}
\label{methods}

\subsection{Problem statement}
Let's assume $\mathbf{y} \in \mathbb{R}^M$ is an $M-$dimensional random vector with non-negative elements, and a true data distribution $p_Y$, which represents the spectral signals. Our goal is to learn an invertible, stable mapping between the approximate data distribution $\hat{p}_Y$ and a latent distribution $p_Z$ (e.g., Gaussian) that will allow for fast evaluation of various inference tasks. However, estimating the full-joint density of very  high-dimensional spectra is a challenging and often intractable task. Therefore, we introduce a second mapping, that transforms $\mathbf{y} \in \mathbb{R}^M$ to $\mathbf{x} \in \mathbb{R}^L$ where $L \ll M$ to discover the spectral latent representation of the signals. Next, we learn an invertible mapping between $\mathbf{x} \sim \hat{p}_X$ (spectral latent variable) and $\mathbf{z} \sim p_Z$ (latent variable). This framework allows us to generate novel samples on the reduced spectral latent space $\mathbf{x} \sim \hat{p}_X$ and use the inverse transformation to map back to the original space $\mathbb{R}^L \rightarrow \mathbb{R}^M$ and therefore approximate the true data distribution.

We also want to estimate an unobserved state vector $\mathbf{v} \in \mathbb{R}^C$ where $C$ is the vector dimensionality. For the ChemCam application we consider, this represents the QQCC, an 8-dimensional vector with the relative weight percentages of 8 major oxides commonly found on Mars.
Given a training dataset of LIBS spectra $\mathbf{y} \in \mathbb{R}^M$ and associated compositions $\mathbf{v} \in \mathbb{R}^C$ (samples generated on Earth), we are interested in constructing a surrogate of the mapping $f: \mathbb{R}^M \rightarrow \mathbb{R}^C$, that will allow us to make predictions of the chemical concentration of novel samples. To calculate uncertainties related to these predictions, we propose an approach based on bootstrapping that allows us to quantify both model and data uncertainties and thus assign measures of accuracy to sample estimates. This approach can be then employed for UQ of data generated by the normalizing flow model.

\subsection{Spectral NMF latent space}
Consider $N$ observations of the random vector $\mathbf{y} \in \mathbb{R}^M$ and let the data matrix be $Y = [\mathbf{y}_1, \mathbf{y}_2, .., \mathbf{y}_N]^T \in \mathbb{R}_{\ge 0}^{N \times M}$. We use non-negative matrix factorization (NMF) to decompose $Y$ into a product of a non-negative basis matrix $X \in \mathbb{R}_{\ge 0}^{N \times L}$ and a non-negative coefficient matrix $V \in \mathbb{R}_{\ge 0}^{L \times M}$, such that $Y \approx XV$ (or equivalently $y_j \approx \sum_{i=1}^{L}x_i V_{ij}$)
\citep{paatero1994positive, wang2012nonnegative}. NMF decomposes each data point into the linear combination of the basis vectors. The NMF optimization problem consists of minimizing the Frobenius norm between $Y$ and $XV$. 


NMF is appropriate for non-negative data and a powerful method for feature selection; and thus allows us to ignore the non-informative LIBS dimensions and enable interpretability of results \citep{rammelkamp2020optimization}.

\subsection{Inference via a spectral normalizing flow}
\label{real-nvp}
We propose the construction of a normalizing flow model on the latent space of the LIBS spectra, obtained by the NMF decomposition, to learn the underlying probability distribution of the spectral latent variable $\mathbf{x} \in \mathbb{R}^L$. The inverse NMF mapping introduced in the previous section can be used to project generated samples back to the physically interpretable space (i.e., $\mathbb{R}^L \rightarrow \mathbb{R}^{M}$).

Normalizing flows are a powerful class of likelihood generative models which transform a base density into a target density by a series of deterministic and invertible transformations \citep{kobyzev2020normalizing}. Consider the base density $p_Z(\mathbf{z})$ (spectral latent variable), the more complex density $p_X(\mathbf{x})$ (latent variable) and an invertible mapping $\mathbf{x} = f(\mathbf{z})$. Under the change of variables formula we can compute the log-likelihood of $\mathbf{x}$ as
\begin{equation}
\label{eq:nf}
    \log (p_X(\mathbf{x}|\theta)) = \log (p_Z(\mathbf{z}|\theta)) + \log  \bigg| \det \bigg( \frac{\partial f(\mathbf{x}|\theta)}{\partial x^T} \bigg) \bigg|
\end{equation}
where $\theta$ represents the trainable parameters of the flow. To train the NF model the negative log-likelihood (NLL) of Eq.\eqref{eq:nf} is minimized.

Here, we parameterize the normalizing flow with a sequence of real-valued non-volume preserving (RealNVP) transformations \citep{dinh2016density}. The RealNVP model, composes two types of invertible transformations: additive coupling layers and rescaling. The model uses the so-called \textit{affine coupling} layers for the coupling flows, which are simple and computationally efficient. The transformation can be written as
\begin{equation}
\label{eq:realnvp}
\begin{aligned}
    \mathbf{x}_{1:d} &= \mathbf{z}_{1:d}  \\
    \mathbf{x}_{d+1:D} &= \mathbf{z}_{d+1:D} \odot \exp (f_{\alpha}(\mathbf{z}_{1:d})) + f_{\mu}(\mathbf{z}_{1:d}),
\end{aligned}
\end{equation}
where $\odot$  is the Hadamard product or element-wise product and the $\exp (\cdot)$ is applied to each element of $\alpha$. The above transformation performs a `1-1' mapping to the first $d$ elements and scales and shifts the remaining $D-d$. By incorporating coupling layers into the flow, the elements are permuted across layers so that a different set of elements is copied each time. Here we model $f_{\alpha}$, $f_{\mu}$ as neural networks. Once the PDF is learned, downstream inference tasks can be performed straightforwardly. In the next Section, we are interested in predicting the elemental composition of novel samples, with their associated uncertainty.

\subsection{Uncertainty quantification via bootsrapping}
\label{uq}

We now aim to construct a mapping between the LIBS signal signatures and QQCC. We train shallow neural networks, one for each oxide element. The models are formed as
\begin{equation}
\label{eq:nn-models}
    v^{(i)} = \sigma(y^T x^{(i)} + b), \quad i=1,..,t
\end{equation}
where $t$ is the total number of oxides to be determined, $w, b$ denote the trainable weights and bias and $\sigma$ is the activation function. We learn the parameters of the models with a training set $\{ (y_i, v_i) \}_{i=1}^{N}$, where $N$ is the total number of samples. The main advantage of such models is that they are both fast to train and result in very good accuracy scores (see Results).

\begin{wrapfigure}{R}{0.4\textwidth}
\centering
\includegraphics[width=0.4\textwidth]{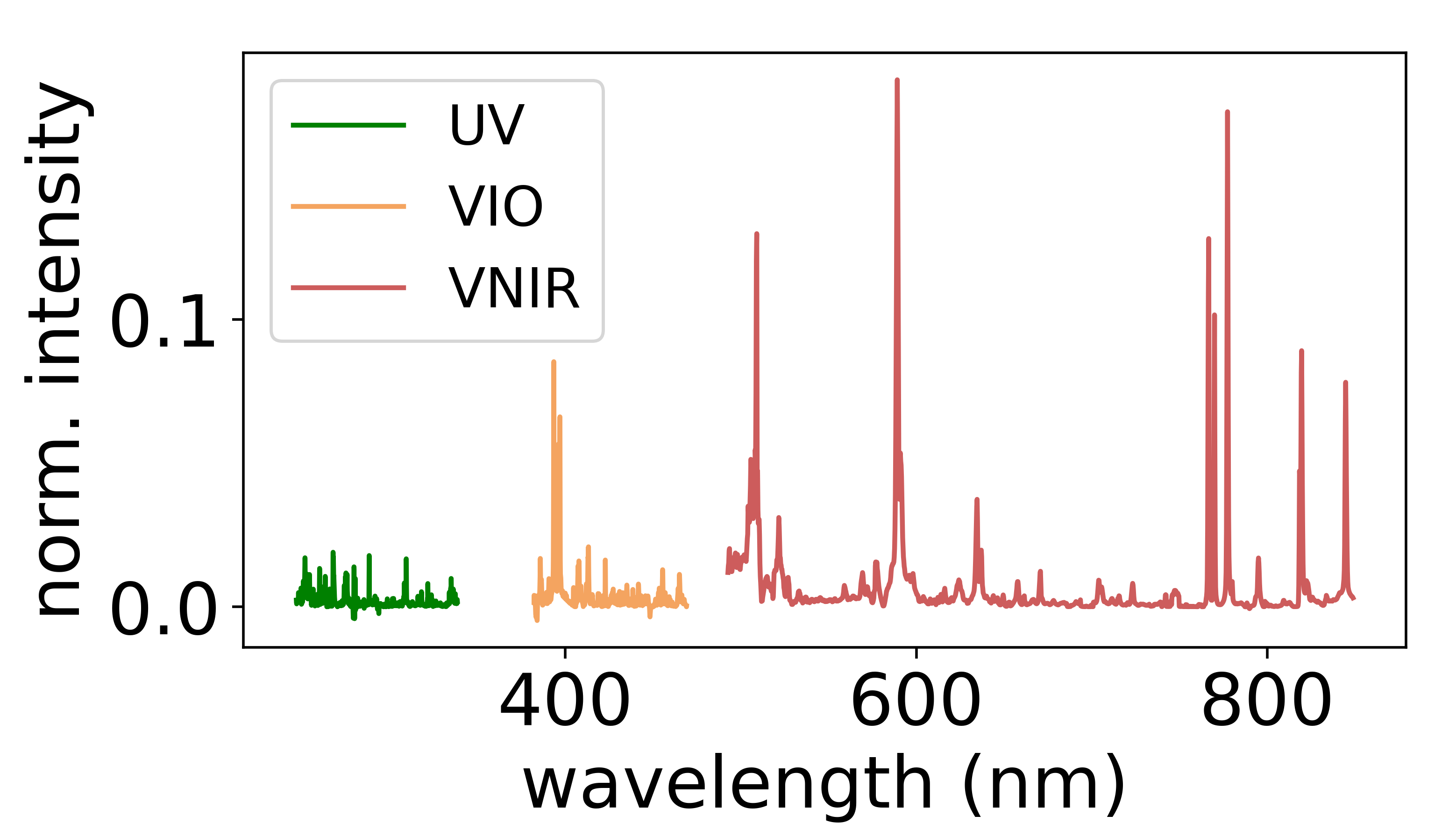}
\caption{\label{fig:random-samples}Random sample generated by the normalizing flow model and transformed to the original space.}
\end{wrapfigure}

To quantify uncertainties related to predictions of elemental compositions we use bootstrapping \citep{kumar2012bootstrap}, a statistics resampling method which allow us to assign measures of accuracy to a sample estimate. In general, bootstrapping performs as well as parametric prediction intervals and its implementation is straightforward. For our application, it does not result in high-computational cost given the choice of simplistic bootstrapped shallow neural networks. In case of more complex models, methods that leverage the last-layer of the network can be employed as they have shown good performance \citep{brosse2020last}. Given a new observation $\mathbf{y}_0 \in \mathbb{R}^{M}$, we can write
\begin{equation}
\label{eq:bootstr}
    \mathbf{v}_0 = \hat{v}_n(\mathbf{y}_0) + r(\mathbf{y}_0)
\end{equation}
where $\hat{v}_n(\mathbf{y}_0)$ represents the model estimate at the $n$-th bootstrap iteration (model uncertainty) and $r(\mathbf{y}_0)$ the predicted residual between true and predicted values which can been modeled for a training dataset with a regression model (data uncertainty). To measure the quality of prediction intervals, we compute the \textit{coverage} of validation samples (the rate at which the actual values fall within the range of the prediction interval). 

\section{Results}

Consider $Y \in \mathbb{R}^{N \times M}$ the LIBS spectra matrix with $N=426$ and $M=5606$. We perform NMF using $L=15$ (selected by 5-fold cross-validation) with transformed data matrix $X \in \mathbb{R}^{N \times L}$. Next we construct a NF model based on the RealNVP architecture with 5 coupling layers and a Gaussian distribution as a base density. We should highlight here the computational advantages of this approach. Constructing a NF model on the 15-dimensional latent space is extremely fast as the training process required less than 1 minute of CPU time. In Figure \ref{fig:random-samples}, we show a novel random sample generated by the NF model which is transformed back to the original space with inverse NMF.
\begin{figure}
    \centering
    \includegraphics[width=0.9\textwidth]{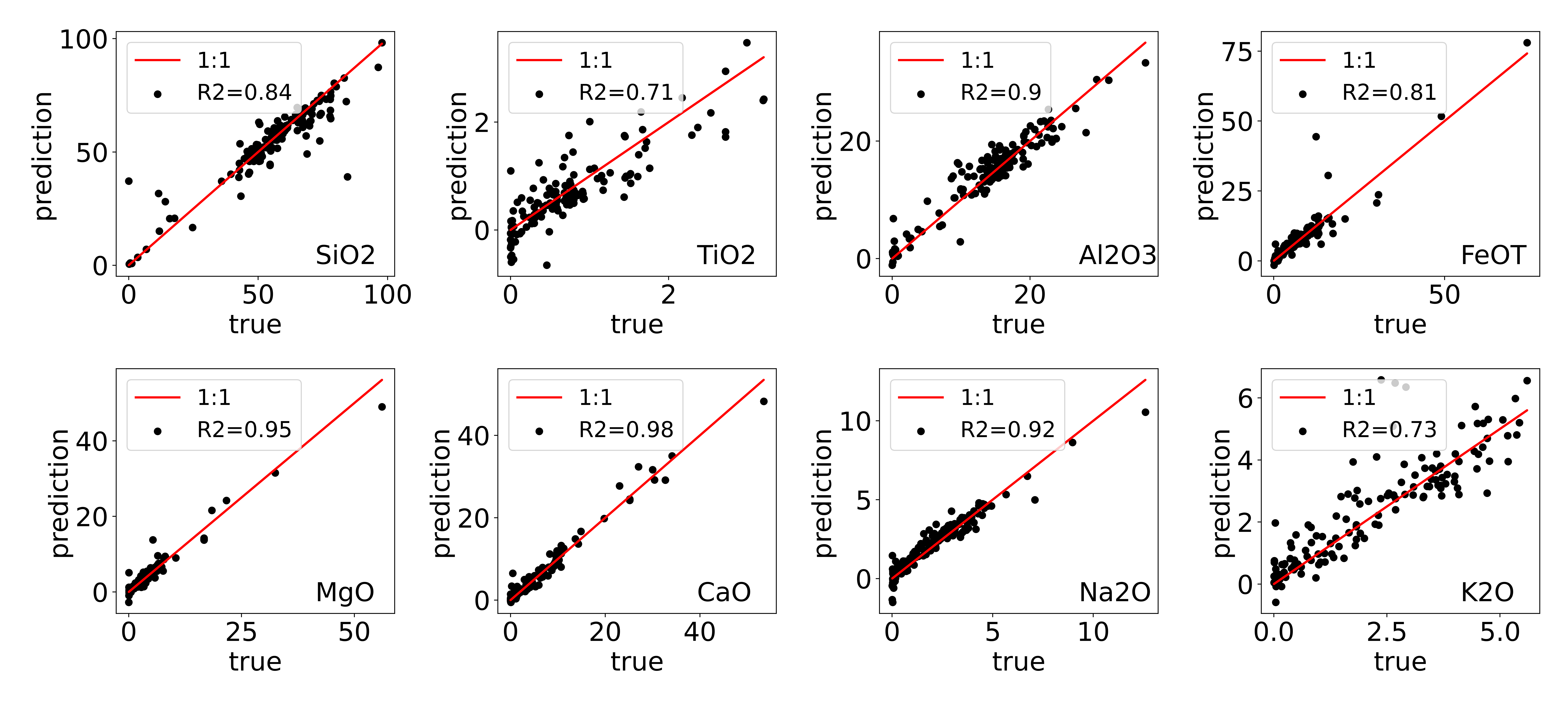}
    \caption{Regression results of the composition of each of the 8 major oxides for a holdout set of 140 samples. Accuracy is measured with the coefficient of determination ($R^2$ score) and x,y axes represent the true and predicted oxide wt.$\%$ respectively.}
    \label{fig:NN-accuracy}
\end{figure}

\begin{table}
\begin{center}
\caption{\label{tab:coverage}Coverage results for 95$\%$ confidence intervals and 140 validation samples.}
\begin{tabular}{ |c|c|c|c|c|c|c|c|c| } 
\hline
oxide & $\text{SiO}_2$ & $\text{TiO}_2$ & $\text{Al}_2\text{O}_3$ & $\text{FeO}_{\text{T}}$ & $\text{MgO}$ & $\text{CaO}$  & $\text{Na}_2\text{O}$ & $\text{K}_2\text{O}$ \\
\hline
coverage ($\%$) & 84.89 & 98.56 & 86.33 & 86.33 & 86.33 & 96.40 & 93.53 & 89.93  \\
\hline
\end{tabular}
\end{center}
\end{table}

We consider 8 oxides and therefore we construct $t=8$, single hidden layer neural network models, with ReLU activation function, trained with stochastic gradient descent (SGD). To measure the accuracy of results we compute the coefficient of determination, calculated as $R^2 = 1-\frac{\text{RSS}}{\text{TSS}}$, where $\text{RSS}$ represents the sum of squares of residuals and $\text{TSS}$ the total sum of squares. We show the accuracy of models in Figure \ref{fig:NN-accuracy} for a holdout set of 140 samples. The results reported show that point estimates are close to the optimal regression (1:1 line) and overall the response is comparable to state-of-the-art deep CNN approaches \citep{castorena2021deep}. For $\text{TiO}_2$ and $\text{K}_2\text{O}$ elements, estimates show larger deviation, which are not considered significant due to the small oxide wt.$\%$ values at these regions. 

\begin{wrapfigure}{R}{0.5\textwidth}
\centering
\includegraphics[width=0.5\textwidth]{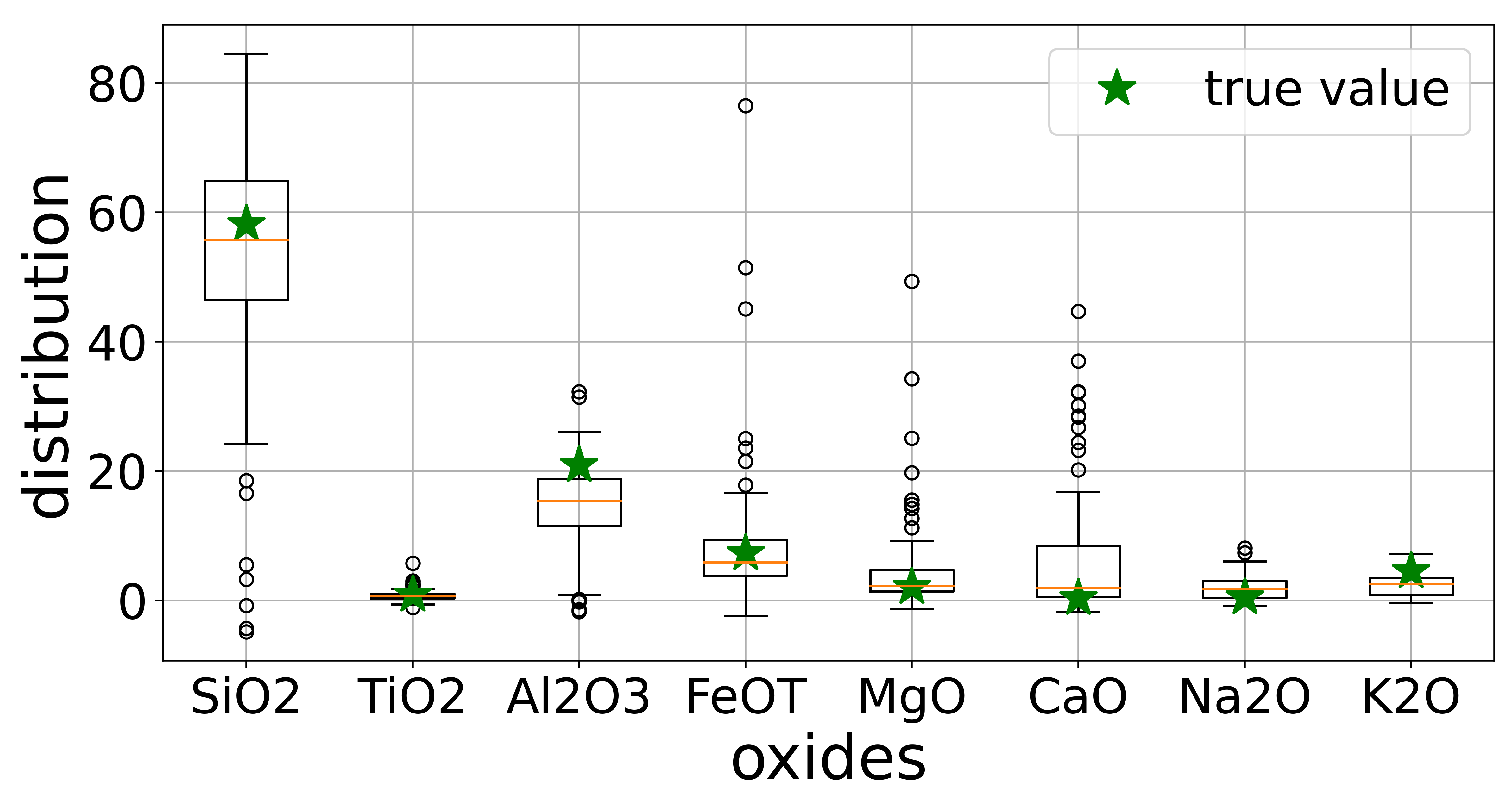}
\caption{\label{fig:bootstrap}Bootstrap results for a random sample.}
\vspace{-5pt}
\end{wrapfigure}

Finally, we perform bootstrapping for a dataset of LIBS spectra samples collected on Earth (with associated ground truth) and we compute the prediction intervals for the same holdout dataset of 140 samples. In Figure \ref{fig:bootstrap}, we plot the distribution of bootstrap predictions with the ground truth for a random Earth sample where box plots represent the first quartile to the third quartile and green stars are the ground truth. Table \ref{tab:coverage} shows the coverage calculated for 95$\%$ prediction intervals for all holdout samples, and we see that the intervals appear to nearly achieve the nominal coverage. The validated approach can be therefore used for making predictions with uncertainty for novel LIBS spectra, generated either by the normalizing flow model or for Martian samples directly collected from ChemCam.

\section{Conclusions}
We showed that the proposed framework provides a straightforward way to perform downstream inference tasks for high-dimensional spectral data by identifying a spectral latent space, estimated as a parsimonious representation of the data, and constructing a spectral normalizing flow model on the reduced space. The proposed approach is ideal for modeling high-dimensional data and enables the learning of complex distributions in a fast and efficient way. Beyond general high-dimensional inference, the proposed UQ approach allows for predictions of state vectors associated with novel out-of-distribution data or data generated by the trained normalizing flow.

\section{Broader impact}
This work provides a robust approach to construct low-dimensional probability densities on spectral latent spaces and quantify uncertainties for predictions related to the elemental compositions of spectral samples generated directly from neural density estimators. Our approach has immediate application for inference and direct UQ of spectral data in different fields such as astronomy, geology, audio signal processing, bioinformatics and more. We believe that this work does not have future societal or ethical consequences.

\acksection
This project was supported by the Laboratory Directed Research and Development program of Los Alamos National Laboratory under project number LDRD-20210043DR. Research was performed while K.K. was an Applied Machine Learning Summer Research Fellow at LANL. 

\small
\bibliography{references}

\providecommand{\latin}[1]{#1}
\makeatletter
\providecommand{\doi}
  {\begingroup\let\do\@makeother\dospecials
  \catcode`\{=1 \catcode`\}=2 \doi@aux}
\providecommand{\doi@aux}[1]{\endgroup\texttt{#1}}
\makeatother
\providecommand*\mcitethebibliography{\thebibliography}
\csname @ifundefined\endcsname{endmcitethebibliography}
  {\let\endmcitethebibliography\endthebibliography}{}
\begin{mcitethebibliography}{11}
\providecommand*\natexlab[1]{#1}
\providecommand*\mciteSetBstSublistMode[1]{}
\providecommand*\mciteSetBstMaxWidthForm[2]{}
\providecommand*\mciteBstWouldAddEndPuncttrue
  {\def\EndOfBibitem{\unskip.}}
\providecommand*\mciteBstWouldAddEndPunctfalse
  {\let\EndOfBibitem\relax}
\providecommand*\mciteSetBstMidEndSepPunct[3]{}
\providecommand*\mciteSetBstSublistLabelBeginEnd[3]{}
\providecommand*\EndOfBibitem{}
\mciteSetBstSublistMode{f}
\mciteSetBstMaxWidthForm{subitem}{(\alph{mcitesubitemcount})}
\mciteSetBstSublistLabelBeginEnd
  {\mcitemaxwidthsubitemform\space}
  {\relax}
  {\relax}

\bibitem[Wiens \latin{et~al.}(2012)Wiens, Maurice, Barraclough, Saccoccio,
  Barkley, Bell, Bender, Bernardin, Blaney, Blank, \latin{et~al.}
  others]{wiens2012chemcam}
Wiens,~R.~C.; Maurice,~S.; Barraclough,~B.; Saccoccio,~M.; Barkley,~W.~C.;
  Bell,~J.~F.; Bender,~S.; Bernardin,~J.; Blaney,~D.; Blank,~J., \latin{et~al.}
   The {ChemCam} instrument suite on the {Mars Science Laboratory (MSL)} rover:
  Body unit and combined system tests. \emph{Space science reviews}
  \textbf{2012}, \emph{170}, 167--227\relax
\mciteBstWouldAddEndPuncttrue
\mciteSetBstMidEndSepPunct{\mcitedefaultmidpunct}
{\mcitedefaultendpunct}{\mcitedefaultseppunct}\relax
\EndOfBibitem
\bibitem[Forni \latin{et~al.}(2013)Forni, Maurice, Gasnault, Wiens, Cousin,
  Clegg, Sirven, and Lasue]{forni2013independent}
Forni,~O.; Maurice,~S.; Gasnault,~O.; Wiens,~R.~C.; Cousin,~A.; Clegg,~S.~M.;
  Sirven,~J.-B.; Lasue,~J. Independent component analysis classification of
  laser induced breakdown spectroscopy spectra. \emph{Spectrochimica Acta Part
  B: Atomic Spectroscopy} \textbf{2013}, \emph{86}, 31--41\relax
\mciteBstWouldAddEndPuncttrue
\mciteSetBstMidEndSepPunct{\mcitedefaultmidpunct}
{\mcitedefaultendpunct}{\mcitedefaultseppunct}\relax
\EndOfBibitem
\bibitem[Boucher \latin{et~al.}(2015)Boucher, Ozanne, Carmosino, Dyar,
  Mahadevan, Breves, Lepore, and Clegg]{boucher2015study}
Boucher,~T.~F.; Ozanne,~M.~V.; Carmosino,~M.~L.; Dyar,~M.~D.; Mahadevan,~S.;
  Breves,~E.~A.; Lepore,~K.~H.; Clegg,~S.~M. A study of machine learning
  regression methods for major elemental analysis of rocks using laser-induced
  breakdown spectroscopy. \emph{Spectrochimica Acta Part B: Atomic
  Spectroscopy} \textbf{2015}, \emph{107}, 1--10\relax
\mciteBstWouldAddEndPuncttrue
\mciteSetBstMidEndSepPunct{\mcitedefaultmidpunct}
{\mcitedefaultendpunct}{\mcitedefaultseppunct}\relax
\EndOfBibitem
\bibitem[Castorena \latin{et~al.}(2021)Castorena, Oyen, Ollila, Legget, and
  Lanza]{castorena2021deep}
Castorena,~J.; Oyen,~D.; Ollila,~A.; Legget,~C.; Lanza,~N. Deep spectral {CNN}
  for laser induced breakdown spectroscopy. \emph{Spectrochimica Acta Part B:
  Atomic Spectroscopy} \textbf{2021}, \emph{178}, 106125\relax
\mciteBstWouldAddEndPuncttrue
\mciteSetBstMidEndSepPunct{\mcitedefaultmidpunct}
{\mcitedefaultendpunct}{\mcitedefaultseppunct}\relax
\EndOfBibitem
\bibitem[Paatero and Tapper(1994)Paatero, and Tapper]{paatero1994positive}
Paatero,~P.; Tapper,~U. Positive matrix factorization: A non-negative factor
  model with optimal utilization of error estimates of data values.
  \emph{Environmetrics} \textbf{1994}, \emph{5}, 111--126\relax
\mciteBstWouldAddEndPuncttrue
\mciteSetBstMidEndSepPunct{\mcitedefaultmidpunct}
{\mcitedefaultendpunct}{\mcitedefaultseppunct}\relax
\EndOfBibitem
\bibitem[Wang and Zhang(2012)Wang, and Zhang]{wang2012nonnegative}
Wang,~Y.-X.; Zhang,~Y.-J. Nonnegative matrix factorization: A comprehensive
  review. \emph{IEEE Transactions on knowledge and data engineering}
  \textbf{2012}, \emph{25}, 1336--1353\relax
\mciteBstWouldAddEndPuncttrue
\mciteSetBstMidEndSepPunct{\mcitedefaultmidpunct}
{\mcitedefaultendpunct}{\mcitedefaultseppunct}\relax
\EndOfBibitem
\bibitem[Rammelkamp \latin{et~al.}(2020)Rammelkamp, Gasnault, Forni, Lasue, and
  Maurice]{rammelkamp2020optimization}
Rammelkamp,~K.; Gasnault,~O.; Forni,~O.; Lasue,~J.; Maurice,~S. Optimization of
  clustering analyses for classification of {ChemCam} data from {Gale}
  {Crater}, {Mars}. European Planetary Science Congress. 2020; pp
  EPSC2020--867\relax
\mciteBstWouldAddEndPuncttrue
\mciteSetBstMidEndSepPunct{\mcitedefaultmidpunct}
{\mcitedefaultendpunct}{\mcitedefaultseppunct}\relax
\EndOfBibitem
\bibitem[Kobyzev \latin{et~al.}(2020)Kobyzev, Prince, and
  Brubaker]{kobyzev2020normalizing}
Kobyzev,~I.; Prince,~S.; Brubaker,~M. Normalizing flows: An introduction and
  review of current methods. \emph{IEEE Transactions on Pattern Analysis and
  Machine Intelligence} \textbf{2020}, \relax
\mciteBstWouldAddEndPunctfalse
\mciteSetBstMidEndSepPunct{\mcitedefaultmidpunct}
{}{\mcitedefaultseppunct}\relax
\EndOfBibitem
\bibitem[Dinh \latin{et~al.}(2016)Dinh, Sohl-Dickstein, and
  Bengio]{dinh2016density}
Dinh,~L.; Sohl-Dickstein,~J.; Bengio,~S. Density estimation using real {NVP}.
  \emph{arXiv preprint arXiv:1605.08803} \textbf{2016}, \relax
\mciteBstWouldAddEndPunctfalse
\mciteSetBstMidEndSepPunct{\mcitedefaultmidpunct}
{}{\mcitedefaultseppunct}\relax
\EndOfBibitem
\bibitem[Kumar and Srivastava(2012)Kumar, and Srivastava]{kumar2012bootstrap}
Kumar,~S.; Srivastava,~A. Bootstrap prediction intervals in non-parametric
  regression with applications to anomaly detection. Proc. 18th ACM SIGKDD
  Conf. Knowl. Discovery Data Mining. 2012\relax
\mciteBstWouldAddEndPuncttrue
\mciteSetBstMidEndSepPunct{\mcitedefaultmidpunct}
{\mcitedefaultendpunct}{\mcitedefaultseppunct}\relax
\EndOfBibitem
\bibitem[Brosse \latin{et~al.}(2020)Brosse, Riquelme, Martin, Gelly, and
  Moulines]{brosse2020last}
Brosse,~N.; Riquelme,~C.; Martin,~A.; Gelly,~S.; Moulines,~{\'E}. On last-layer
  algorithms for classification: Decoupling representation from uncertainty
  estimation. \emph{arXiv preprint arXiv:2001.08049} \textbf{2020}, \relax
\mciteBstWouldAddEndPunctfalse
\mciteSetBstMidEndSepPunct{\mcitedefaultmidpunct}
{}{\mcitedefaultseppunct}\relax
\EndOfBibitem
\end{mcitethebibliography}



\end{document}